\documentclass{article} 
\usepackage{graphicx}
\usepackage{floatrow}
\usepackage{nips13submit_e,times}
\usepackage{hyperref}
\usepackage{url}

\title{Deep Belief Networks for Image Denoising}

\author{
Mohammad Ali Keyvanrad\\
Dept. of Computer Engineering and IT,\\
Amirkabir University of Technology,\\
Tehran, Iran\\
\texttt{keyvanrad@aut.ac.ir} \\
\And
Mohammad Pezeshki\\
Dept. of Computer Engineering and IT,\\
Amirkabir University of Technology,\\
Tehran, Iran\\
\texttt{m.pezeshki@aut.ac.ir} \\
\AND
Mohammad Mehdi Homayounpour\\
Dept. of Computer Engineering and IT,\\
Amirkabir University of Technology,\\
Tehran, Iran\\
\texttt{homayoun@aut.ac.ir} \\
}

\nipsfinalcopy 

\begin{document}

\maketitle

\begin{abstract}
Deep Belief Networks which are hierarchical generative models are effective tools for feature representation and extraction. Furthermore, DBNs can be used in numerous aspects of Machine Learning such as image denoising. In this paper, we propose a novel method for image denoising which relies on the DBNs' ability in feature representation. This work is based upon learning of the noise behavior. Generally, features which are extracted using DBNs are presented as the values of the last layer nodes. 
We train a DBN a way that the network totally distinguishes between nodes presenting noise and nodes presenting image content in the last later of DBN, i.e. the nodes in the last layer of trained DBN are divided into two distinct groups of nodes. 
After detecting the nodes which are presenting the noise, we are able to make the noise nodes inactive and reconstruct a noiseless image.
In section 4 we explore the results of applying this method on the MNIST dataset of handwritten digits which is corrupted with \textit{additive white Gaussian noise (AWGN)}. A reduction of 65.9\% in average \textit{mean square error (MSE)} was achieved when the proposed method was used for the reconstruction of the noisy images.
\end{abstract}

\section{Introduction}
Image signals are often corrupted due to noise. Removing noise from image is an important issue in computer vision, because this step could be the preprocessing step of many other applications.
Up to now various methods have been proposed to remove noise (denoise) from visual data. Focus of many of these methods is on Fourier Analysis [1], Spatial Filtering [2], and Wavelet Transform [3]. Also there are some other methods based on Spare Coding and Dictionary Learning [4]. On the other hand, Machine Learning tools such as Convolutional Neural Networks (CNN) or Deep Neural Networks (DNN) have been used in several papers to tackle this issue[5][6]. One of the biggest difficulties in training such deep networks is that the cost function of such deep architectures gets stuck in poor local optima due to random initialization of weights. Hinton et al. [7] proposed a new greedy layer-wise algorithm to tackle this issue and introduced Deep Belief Networks (DBNs).
DBNs are able to present a good "feature" representation of data. These features which are defined as the properties of input data are presented as nodes of the last layer of DBN.  As a result, in this paper we train a DBN such that it learns to extract image features. The trained DBN distinguishes between "noise features" and "clean image features" in the last layer and presents them into two distinct groups of nodes. Furthermore, DBNs are capable of reconstructing the input data based on the values of last layer nodes. Subsequently, if we eliminate the effects of nodes presenting the noise, the reconstructed image will be noiseless.
From now on,we called the nodes presenting noise and the nodes presenting image content ``noise nodes" and ``image nodes", respectively.
The rest of the paper is organized as follows: In Section 2 we briefly describe Deep Learning. In Section 3 we describe the learning process. Section 4 is our experimental results and finally, we conclude the paper in Section 5.
\section{Deep Learning}
Restricted Boltzmann Machines (RBMs) are the building blocks of DBNs. Hence, in this section first we briefly describe RBMs and then will explore DBNs.

\subsection{Restricted Boltzmann Machines}
\begin{figure}[h]
\centering
\includegraphics[height=3.0cm]{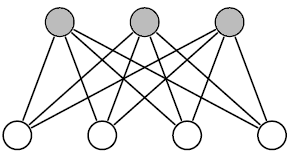}
\caption{Structure of a Restricted Boltzmann Machine.}
\label{RBM}
\end{figure}

Boltzmann Machines (BMs) and Restricted Boltzmann Machines (RBMs) were introduced in 1980s. But they attracted more attention since 2006 after Hinton et al. paper [8]. He showed that a very powerful neural network can be made by stacking RBMs.
RBMs are a kind of Markov Random Fields (MRF) which have a two-layer structure. One layer is called visible and another is called hidden layer. They are restricted to have no visible-visible or hidden-hidden connection and connections are inter layer. A graphical depiction of RBM is shown in Figure \ref{RBM}.

A joint configuration, 
\begin{math}
(v,h)
\end{math}
 of visible and hidden units has an energy given by [9]:
\begin{equation}
E(v,h) = -{\sum_{i\ \in visible}}a_{i}v_{i} - {\sum_{j\ \in hidden}}b_{j}h_{j} - {\sum_{i,j}}v_{i}h_{j}w_{i,j} 
\end{equation}

where 
\begin{math}
v_{i}, h_{j}
\end{math}
 are the binary states of visible unit 
 \begin{math}
 i
 \end{math}
  and hidden unit 
 \begin{math}
 j
 \end{math}
 and
 \begin{math}
 a_{i},b_{j}
 \end{math} 
   are their biases and 
 \begin{math}
 w_{ij}
 \end{math}   
    is the weight between them. The network assigns a probability to every possible pair of a visible and a hidden vector via this energy function [10]:
\begin{equation}
p(v,h) = \frac{e^{-E(v,h)}}{\sum_{v,h} e^{-E(v,h)}}
\end{equation}

The probability that the network assigns to a visible vector, 
\begin{math}
v
\end{math}
, is given by summing over all possible hidden vectors:
\begin{equation}
p(v) = \frac{\sum_{h}e^{-E(v,h)}}{\sum_{v,h} e^{-E(v,h)}}
\end{equation}

\begin{equation}
\frac{\partial \log p(v))}{\partial \omega _{i,j}} = <v_{i}h_{j}>_{data} - <v_{i}h_{j}>_{model }
\end{equation}

where the angle brackets are used to denote expectations under the distribution species by the subscript that follows. This leads to a very simple learning rule for performing stochastic steepest ascent in the log probability of the training data:
\begin{equation}
\Delta \omega _{ij} = \epsilon (<v_{i}h_{j}>_{data} - <v_{i}h_{j}>_{model })
\end{equation}

where 
\begin{math}
\epsilon
\end{math} 
is a learning rate.
But, exact maximum likelihood learning in this model is intractable because exact computation of the expectation model is very expensive. Hence, in practice, learning is done by following an approximation to the gradient of a different objective function, called the \textit{“Contrastive Divergence”} (CD) [11].






\subsection{Deep Belief Networks}

\begin{figure}[h]
\centering
\includegraphics[height=7.0cm]{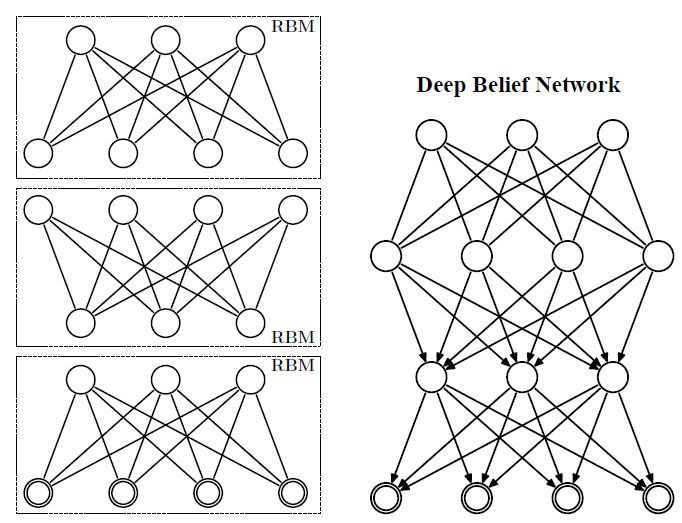}
\caption{\textbf{Left}: Greedy learning a stack of RBM’s in which the samples from the lower-level RBM are used as the data for training the next RBM. \textbf{Right}: The corresponding Deep Belief Network [13].}
\label{DBN}
\end{figure}

One of the main problems in training deep networks is how to initialize weights. It is difficult to optimize the weights in nonlinear Deep Networks with multiple hidden layers. With good initial weights, gradient descent works well, but finding such initial weights requires a very different type of algorithm that learns one layer of features at a time. Hinton et al. [1] introduced a new algorithm to solve the above problem based on the training of a sequence of RBMs. 
To construct a DBN we train sequentially as many RBMs as the number of hidden layers in the DBN, i.e. for a DBN with \textit{h} hidden layers we have to train \textit{h} RBMs. These RBMs are placed one on top of the other. Figure \ref{DBN} gives an overview of the basic concept.
For the first RBM, which consists of the DBN’s input layer and the first hidden layer, the input of RBM is the training set. For the second RBM, which consists of the DBN’s first and second hidden layers, the input is the output of the previous RBM and so for other RBMs. 

After performing this layer-wise algorithm we have obtained a good initialization for the hidden weights and biases of the DBN and then the DBN is fine-tuned with respect to a typical supervised criterion (such as \textit{mean square error} or \textit{cross-entropy}) .




\section{Learning}
\label{learning}

\begin{figure}[h]
\floatbox[{\capbeside\thisfloatsetup{capbesideposition={right,bottom},capbesidewidth=4.5cm}}]{figure}[\FBwidth]
{\caption{\textit{Relative activity} of the last layer nodes can be computed by subtracting the last nodes values constructed by a noiseless image and its corresponding noisy image.}}
{\includegraphics[width=5.7cm]{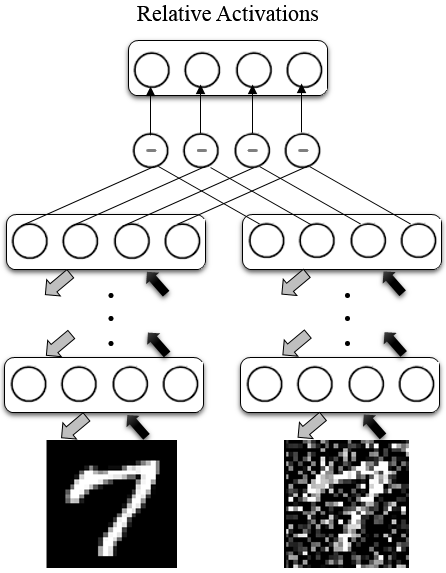}}
\label{RA}
\end{figure}

To train a DBN for image denoising, the normalized values of an image pixels are used. Using min-max normalization, grayscale value of a pixel (an integer between 0 and 255) is transformed to a floating point number between 0 and 1. Unlike first and last layer of DBN, other layers have binary nodes.
The main idea is to train a DBN such that it learns to map noisy images to images with lower noise or even without noise. The idea can be implemented by learning the behavior of noise and image contents and presenting these behaviors in some nodes at the last layer of the network.
The network is trained with a collection consisted of both noisy and noiseless  images. We used a criterion called \textit{relative activity} to detect noise nodes. \textit{Relative activity} of each node is defined as the difference between two values of a particular node resulted from feeding the network using a noiseless image and its corresponding noisy image (Two images with same contents but one of them is corrupted by noise). As a result, if a particular node is a noise node, it should have higher \textit{relative activity}. On the other hand, if it is an image node, it should have lower \textit{relative activity}. This theory is justified by the fact that the activation of image nodes should be same for both noiseless and its corresponding noisy images. This process is illustrated in Figure ~\ref{RA}.

By performing above operation for all images and averaging on the values of each node in the last layer, \textit{average relative activity} of the last layer nodes is computed. The nodes that still have high \textit{average relative activity} are considered noise nodes.
Now that the noise nodes are discovered, the next step is to lower their activity. Since noise nodes do not change much when clean images are fed to the network, we choose their average value for all clean images as their neutral values (the values which make nodes inactive).
Finally, the noise nodes are inactive and consequently, a noiseless image can be reconstructed as it is shown in Figure 4.

\begin{figure}[h]
\floatbox[{\capbeside\thisfloatsetup{capbesideposition={right,bottom},capbesidewidth=4.5cm}}]{figure}[\FBwidth]
{\caption{Nodes in the last layer of the DBN are presenting the noise and the contents of the input image. By reducing the noisy nodes (gray nodes in the figure above) activity, a noiseless image will be created.  Right side image is the noisy image and the left side image is noiseless reconstruction.}}
{\includegraphics[width=6.9cm]{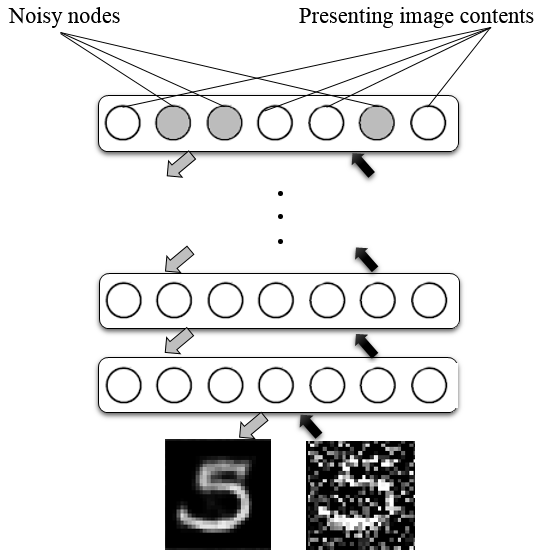}}
\label{IL}
\end{figure}

\section{Experimental Results}
\label{experiments}

MNIST dataset is a dataset of handwritten digits consisted of 60,000 images for training and 10,000 images for test. To model natural noise, we added \textit{additive white Gaussian noise (AWGN)} to images with a variance of 0.20. Therefore, our new dataset was consisted of 120,000 noisy and clean images along with 10,000 noisy images for test (the whole test set is noisy). We used a subset of training set with 20,000 elements for the training phase and the whole test set for the test phase. According to empirical results we created a DBN with 4 hidden layers: 784-1000-500-250-100. We trained this network by 200 batches of data each including 100 images.

According to previous discussions we used \textit{“relative activity”} to find noise nodes in the last layer of the DBN:
For all images in the dataset, we put a clean image and then its corresponding noisy image as the input of the network. Afterward, we computed the difference between the last nodes' values. The average of difference in each node considering all images showed \textit{“average relative activity”} of nodes (noisy vs. clean). Based on experimental results, nodes with an \textit{“average relative activity”} higher than 0.9 were considered noise nodes.

\begin{figure}[h]
\floatbox[{\capbeside\thisfloatsetup{capbesideposition={right,bottom},capbesidewidth=4.5cm}}]{figure}[\FBwidth]
{\caption{From \textbf{left} to \textbf{right}: Noiseless images, Noisy images (AWGN with 0.20 variance), Reconstruction without eliminating any noisy node, Reconstruction with eliminating noise nodes. As it is clear, the reconstructed results have much lower noise after eliminating noise nodes.}\label{fig:test}}
{\includegraphics[width=5.7cm]{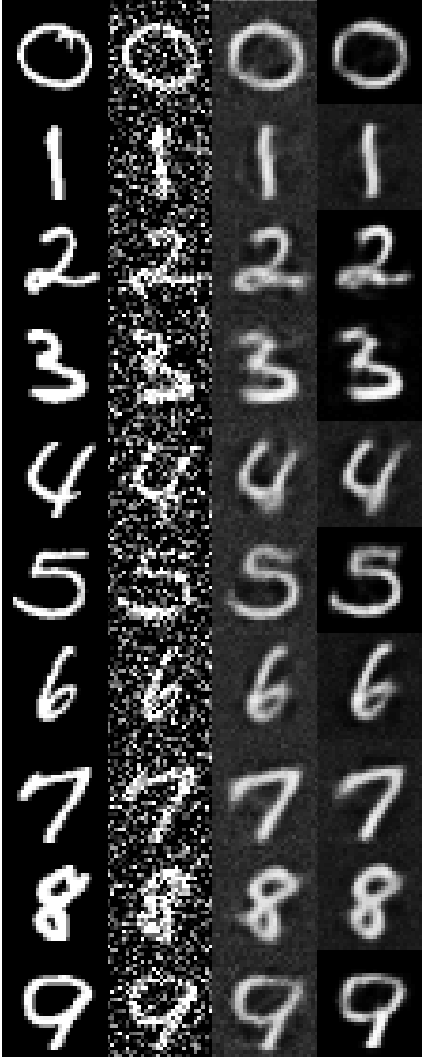}}
\label{RES}
\end{figure}

Now for all 10000 clean images in our training set, we compute the values of nodes in the last layer of our trained DBN. The average of these values for each node considered neutral value of node. 
Finally, to reconstruct a noiseless image from a noisy image, we change the values noise nodes to their neutral values. As a result, noise nodes are inactive and reconstruction would be noiseless. Figure ~\ref{RES} shows how the reconstructed results have lower noise. Also Table ~\ref{TRES} shows that a reduction of 65.9\% in average Mean Square Error (MSE) was achieved when the proposed method was used for the reconstruction of the noisy images.

\section{Conclusion and Future Works}
\label{conclusion}
In this paper, a novel method for image denoising was proposed. The proposed method makes a model to learn noise behavior using a Deep Belief Network (DBN) and tries to present noise and image contents behavior into two distinct groups of nodes. Then by omitting noisy nodes, the network will be able to produce a noiseless (clean) image by reconstruction of the input image. In our work, thresholds for detection of noise nodes were determined manually. 
Future works will include:
\\1.	Using an automatic technique to determine thresholds for detecting the noisy nodes in denoising technique presented in this paper. 
\\2.	Employing our denoising approach to tackle some other issues in Computer Vision, Speech Recognition, etc.
These areas will be addressed in future phases of this project.

\begin{table} [h]
\begin{center}
  \begin{tabular}{ l  || r }
    \hline
    \textbf{Data} & \textbf{Mean Square Error (MSE)} \\ \hline
    Noisy Image & 0.0966 \\ \hline
    Reconstruction without eliminating any node & 0.0416 \\ \hline
    Reconstruction with eliminating noise node & \textbf{0.0329} \\    
    \hline
  \end{tabular}      
  \caption {The table above shows the average MSE for different kinds of reconstructions. A reduction of \textbf{65.9\% (0.0966 to 0.0329)} in average MSE was achieved by reconstruction after eliminating noisy nodes.}
\end{center}
\label{TRES}
\end{table}

\section{References}

\small{
[1]	Brigham, E. O., \& Morrow, R. E. (1967). The fast Fourier transform. Spectrum, IEEE, 4(12), 63-70.

[2]	Kervrann, C., \& Boulanger, J. (2006). Optimal spatial adaptation for patch-based image denoising. Image Processing, IEEE Transactions on, 15(10), 2866-2878.

[3]	Pan, Q., Zhang, L., Dai, G., \& Zhang, H. (1999). Two denoising methods by wavelet transform. Signal Processing, IEEE Transactions on, 47(12), 3401-3406.

[4]	Elad, M., \& Aharon, M. (2006). Image denoising via sparse and redundant representations over learned dictionaries. Image Processing, IEEE Transactions on, 15(12), 3736-3745.

[5]	LeCun, Y., Kavukcuoglu, K., \& Farabet, C. (2010, May). Convolutional networks and applications in vision. In Circuits and Systems (ISCAS), Proceedings of 2010 IEEE International Symposium on (pp. 253-256). IEEE.

[6]	Xie, Junyuan, Linli Xu, and Enhong Chen. "Image denoising and inpainting with deep neural networks." Advances in Neural Information Processing Systems. 2012.

[7]	Hinton, G. E., \& Salakhutdinov, R. R. (2006). Reducing the dimensionality of data with neural networks. Science, 313(5786), 504-507.

[8]	Hinton, G. E., Osindero, S., \& Teh, Y. W. (2006). A fast learning algorithm for deep belief nets. Neural computation, 18(7), 1527-1554.

[9]	Hinton, G. E. (2002). Training products of experts by minimizing contrastive divergence. Neural computation, 14(8), 1771-1800.

[9]	Vincent, P., Larochelle, H., Lajoie, I., Bengio, Y., \& Manzagol, P. A. (2010). Stacked denoising autoencoders: Learning useful representations in a deep network with a local denoising criterion. The Journal of Machine Learning Research, 9999, 3371-3408.

[10]	Hinton, G. (2010). A practical guide to training restricted Boltzmann machines. Momentum, 9(1).

[11]	Salakhutdinov, R. (2009). Learning deep generative models (Doctoral dissertation, University of Toronto).

}

\end{document}